\documentclass[letterpaper, 10 pt, conference]{ieeeconf} 

\IEEEoverridecommandlockouts                              

\usepackage{graphics} 
\usepackage{epsfig} 
\usepackage{mathptmx} 
\usepackage{amsmath} 
\usepackage{color}

\title{\LARGE \bf
Geometry-Based Next Frame Prediction from Monocular Video 
}

\author{Reza Mahjourian$^{1*}$ \quad Martin Wicke$^{2}$\quad Anelia Angelova$^{2}$
\thanks{*This work was done while at Google Brain.}
\thanks{$^{1}$Department of Computer Science, University of Texas at Austin, Austin, TX 78712, USA
        {\tt\small reza@cs.utexas.edu}}%
\thanks{$^{2}$Google Brain, 
        {\tt\small wicke@google.com,anelia@google.com}}%
}

\newcommand{\ie}[1][ ]{{\em i.\thinspace e\@.}#1}
\newcommand{\eg}[1][ ]{{\em e.\thinspace g\@.}#1}

\begin{document}

\maketitle
\thispagestyle{empty}
\pagestyle{empty}

\begin{abstract}

We consider the problem of next frame prediction from video input. A recurrent convolutional neural network is trained to predict depth from monocular video input, which, along with the current video image and the camera trajectory, can then be used to compute the next frame. Unlike prior next-frame prediction approaches, we take advantage of the scene geometry and use the predicted depth for generating the next frame prediction. Our approach can produce rich next frame predictions which include depth information attached to each pixel.  Another novel aspect of our approach is that it predicts depth from a sequence of images (e.g. in a video), rather than from a single still image.

We evaluate the proposed approach on the KITTI dataset, a standard dataset for benchmarking tasks relevant to autonomous driving.  The proposed method produces results which are visually and numerically superior to existing methods that directly predict the next frame.  We show that the accuracy of depth prediction improves as more prior frames are considered.  

\end{abstract}

\section{INTRODUCTION}

Scene understanding, \ie[,] attaching meaning to images or video, is a problem with many potential applications in computer vision, computer graphics, and robotics. We are interested in a particular test for such approaches: whether they are able to predict what happens next, \ie[,] given a video stream, predict the next frame in the video sequence.

\begin{figure}
\centering
\includegraphics[width=0.47\textwidth]{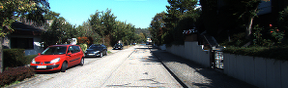}
\includegraphics[width=0.47\textwidth]{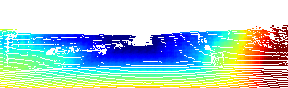}
\includegraphics[width=0.47\textwidth]{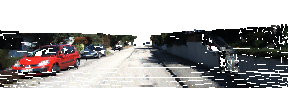}
\includegraphics[width=0.47\textwidth]{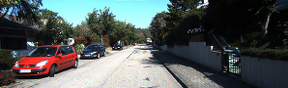}
\caption{Sample prediction produced by our method.  From top to bottom: 1) Input frame (e.g. from a sequence). 2) Depth prediction. 3) Next frame prediction.  4) Ground truth next frame.  Note that the car has moved forward which is accurately recovered by our algorithm.}  
\label{fig:sample_prediction}
\end{figure}

Traditionally, predictive approaches have been model-based, with strong assumptions about what kind of scenes are permissible~\cite{michalski2014modeling,fragkiadaki2016learning} \eg[,] a bouncing ball or a rigid object. Such assumptions lead to a parametric model of the world, which can be fitted to the observations. For example, assuming that a camera observes a single object, one can conceivably fit the degrees of freedom of the object and their rates of change to best match the observations. Then, one can use generative computer graphics to predict the next frame to be observed. While model-based methods~\cite{michalski2014modeling,fragkiadaki2016learning} perform well in restricted scenarios, they are not suitable for unconstrained environments. Model-free approaches, on the other hand, do not rely on any assumption about the world and predict future frames simply based on the video stream~\cite{mathieu2015deep}. The simplest such techniques use a 2D optical flow field computed from the video to warp the last frame~\cite{walker2015dense}. The resulting next frame prediction is not optimized for visual quality, but works well in some applications (e.g. video compression). 

We propose a \textit{geometry-based} next frame prediction model, which learns to predict a depth map of the scene from \textit{a sequence} of previous RGB video frames as input.
Similar to classic model-based approaches, we then use generative computer graphics to render the next video frame using our predicted depth map, the current video frame, and the camera trajectory---which can be obtained from inertial measurements or from GPS.  This is in contrast to previous next frame prediction methods, which predict unstructured RGB values~\cite{mathieu2015deep,finn2016unsupervised}.  Using the scene geometry allows our method to produce more accurate and realistic next frame renderings (Figure~\ref{fig:sample_prediction}).  We are not aware of other approaches that use depth for next frame predictions.

\begin{figure}
\centering
\includegraphics[width=0.47\textwidth]{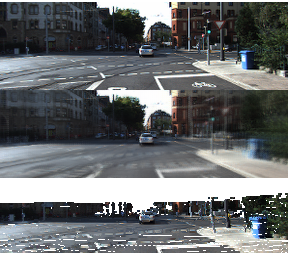}
\caption{Comparison of next frame predictions generated by our approach (bottom) with  prior work~\cite{finn2016unsupervised} (middle).  The input frame is at the top.  Trying to predict the RGB values, as it is commonly done, results in a blurry image (middle).}
\label{fig:comp}
\end{figure}

While our approach is inherently model-based, it produces one of the most general models possible: a depth map for the scene from video input.  This type of model has the advantage that it does not impose any assumptions on the scene and therefore does not limit its generality.  To generate depth predictions we propose a model-free method based on a recurrent convolutional neural network (RCNN), which consists of convolutional LSTM units~\cite{shi2015convolutional} instead of fully-connected units~\cite{hochreiter2016long}.  The LSTM units have the ability to take into account not only the current frame, but a history of video frames of theoretically unbounded length.  While our experiments focus on predicting the next frame, extension to multiple future frames is natural:  Depth predictions can be produced for multiple frames ahead.  Similarly, the camera's near-future trajectory can be predicted from prior observations.  This gives a powerful approach for future frame modeling.

Recent frame prediction methods based on neural networks \cite{ranzato2014video,mathieu2015deep,oh2015action,finn2016unsupervised} train a network to predict the next frame directly from the video stream.  These methods typically use a loss function based on the raw RGB values, which results in blurry predictions (Figure~\ref{fig:comp}), especially for scenes with large motions.  None of these approaches utilize the geometry of the scene.

Depth estimation is an important problem for scene understanding. Recent learning-based approaches~\cite{eigen2015predicting,rao2016monocular3d} propose to predict depth from a single RGB image, whereas classic geometry-based methods~\cite{becker2013variational} use multiple frames. Our approach incorporates the benefits of both by proposing a generic predictive model which utilizes multiple input frames. Our experiments show that using a sequence of frames improves the accuracy of depth predictions.

Our method produces rich next frame predictions which include depth information attached to each pixel.  This 3D representation is more suitable for predicting the scene as a result of the viewer's own motion (ego-motion). For example, it can be used to generate hypothetical next frame predictions as a result of an exploratory or hypothetical action.  

We evaluate our approach on the KITTI raw dataset~\cite{geiger2013vision}. The dataset includes stereo video, 3D point clouds for each frame, and the vehicle trajectory.  We only use monocular video, and show that we can extract a depth stream from the monocular video and predict the next video frame with high accuracy.  We show that this yields better outcomes in terms of visual quality, as well as, quantitative metrics, namely Peak Signal to Noise Ratio (PSNR) and the Structural Similarity Index Measure (SSIM)~\cite{hore2010image}.

\begin{figure*}
\centering
\includegraphics[width=0.7\textwidth,trim={0 0.2cm 0 0.2cm}]{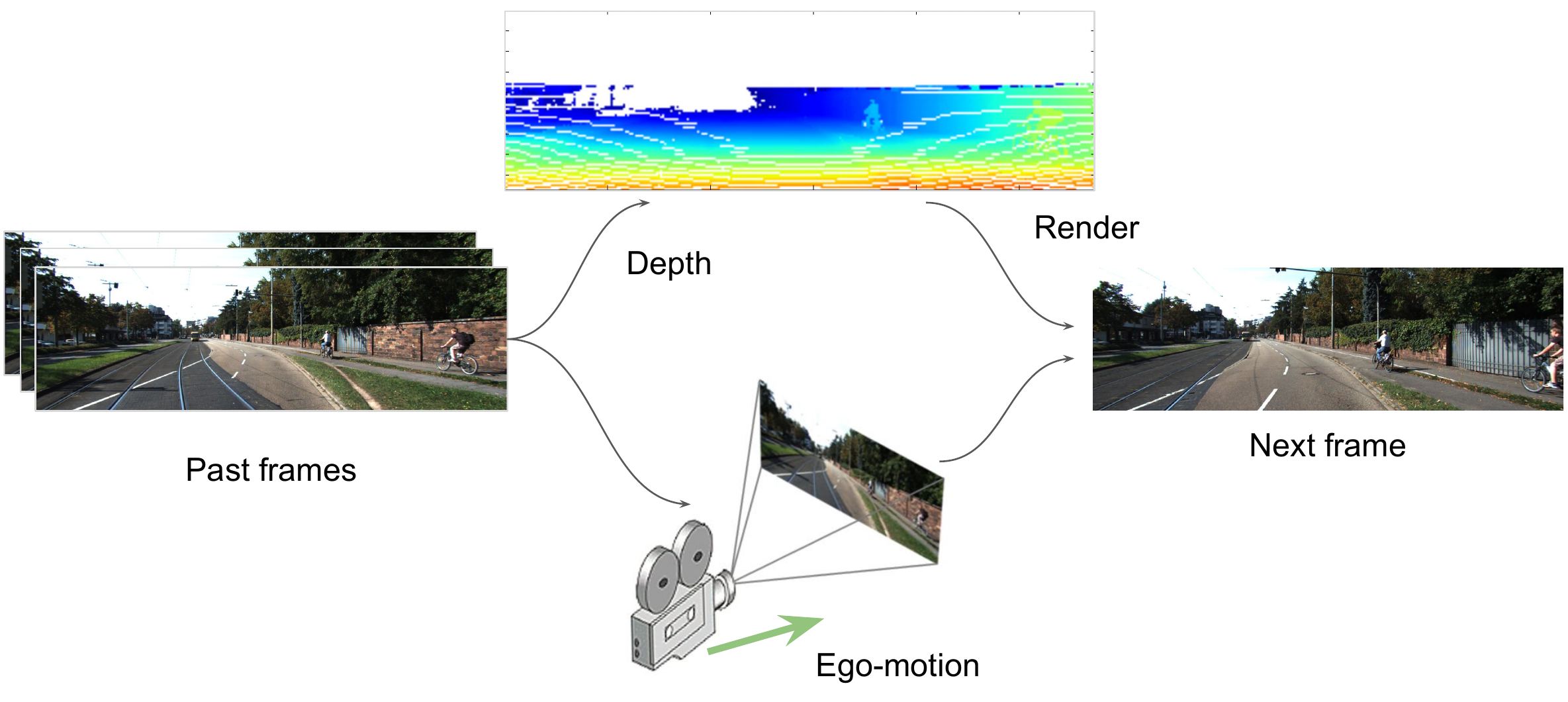}
\caption{Overview of the proposed method: A recurrent convolutional neural network is trained to predict depth from monocular video input, which, along with the current video image and the camera trajectory, can then be used to compute the next frame.}
\label{fig:approach}
\end{figure*}

The main contributions of this work are:
\begin{itemize}
\item We propose a recurrent neural network architecture to predict depth from a sequence of monocular video frames.  The network uses convolutional LSTM units to capture the motion of the scene between subsequent frames.  Based on the motion patterns for different regions in the source image, the network can produce a better estimate on the scene depth.

\item We propose a model for generating next frame predictions based on the geometry of the scene as captured in a depth map, and the camera's trajectory.  The rich 3D representation can also be used to simulate potential next frames as a result of hypothetical ego-motions.
\end{itemize}

To summarize the advantages of our proposed next frame prediction model: 1) Using a convolutional-LSTM architecture allows the model to work with an arbitrary number of input frames.  2) Incorporating more frames improves the accuracy of depth predictions, as evidenced by our experiments.  3) Purely geometry-based approaches for scene modeling or reconstruction do not have predictive capabilities, whereas neural networks can generalize to future frames.  4) Incorporating the scene geometry in the model allows for producing next frame predictions with significantly higher quality and sharpness.  5) Producing rich 3D predictions is beneficial to any decision-making algorithm feeding on the outputs of the model.


\section{RELATED WORK}

Scene understanding~\cite{geiger20143d} is a central topic in computer vision with problems including object detection~\cite{dollar2012pedestrian,benenson2014ten,bahlmann2005asystem}, tracking~\cite{caraffi2012asystem,petrich2016mapbased}, segmentation~\cite{alvarez2012road,schneider2016semanticstixels}, and scene reconstruction~\cite{rao2016monocular3d}.


A few methods~\cite{eigen2015predicting,chen2016single,wang2015towards,laina2016deeper,garg2016unsupervised} have demonstrated learning depth from a single image using deep neural networks. Eigen and Fergus~\cite{eigen2015predicting} use a multi-scale setup to predict depth at multiple resolutions, whereas~\cite{laina2016deeper} uses deeper models to improve the quality of predictions. There are also pure geometry-based approaches~\cite{becker2013variational} that estimate depth from multiple images. Similarly, our approach uses a sequence of images for better depth estimation, but in a learning-based setting.


Unsupervised learning from large unlabeled video datasets has been a topic of recent interest~\cite{srivastava2015unsupervised,vondrickanticipating,oh2015action,mathieu2015deep}. The works in~\cite{ranzato2014video,mathieu2015deep,xue2016probabilistic,finn2016unsupervised} use neural networks for next frame prediction in video.  These methods typically use a loss function based on the RGB values of the pixels in the predicted image.  This results in conservative and blurry predictions where the pixel values are close to the target values, but rarely identical to them.  In contrast, our proposed method produces images whose RGB distribution is very close to the target next frame.  Such an output is more suitable for detecting anomalies or surprising outcomes where the predicted next frame does not match the future state.

\section{NEXT FRAME PREDICTION METHOD}


The problem that our proposed method addresses can be defined as follows.  Given a sequence of RGB frames $\{X_1, X_2, \ldots, X_{k-1}\}$, and a sequence of camera poses $\{P_1, P_2, \ldots P_k\}$, predict the next RGB frame $X_k$.


Our proposed method predicts two depth maps $D_{k-1}$ and $D_k$ corresponding to frames $k-1$ and $k$.  The depth map $D_{k-1}$ is predicted directly from the sequence of images $X_1 \ldots X_{k-1}$.  The depth map $D_k$ is constructed from $D_{k-1}$ and the camera's ego-motion from $P_{k-1}$ to $P_k$ (Figure~\ref{fig:approach}).

The next frame prediction $X_k$ is constructed from the RGB frame $X_{k-1}$ and the two depth maps $D_{k-1}, D_k$ using geometric projections and transformations.

\begin{figure}
\centering
\includegraphics[width=0.5\textwidth]{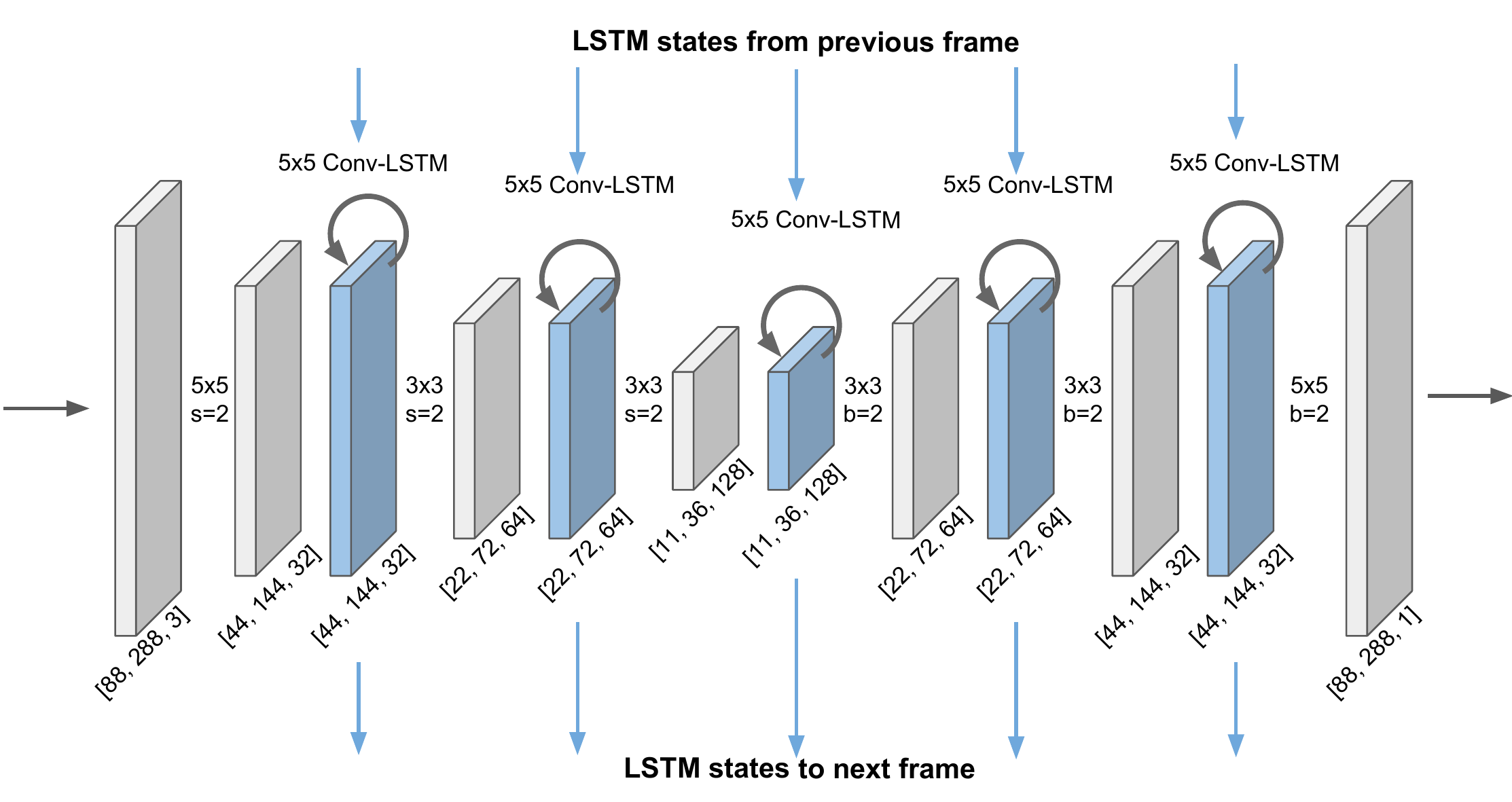}
\caption{The depth prediction neural network using convolutional LSTM cells.  The model receives a sequence of RGB images, each with size $88 \times 288 \times 3$.  It produces depth predictions of size $88 \times 288 \times 1$.  The encoder uses convolutions with stride two to downsize the feature maps.  The decoder uses depth-to-space layers with block size two followed by convolutions with stride one to upsize the feature maps.}
\label{fig:model}
\end{figure}

\begin{figure*}
\centering
\includegraphics[width=0.71\textwidth]{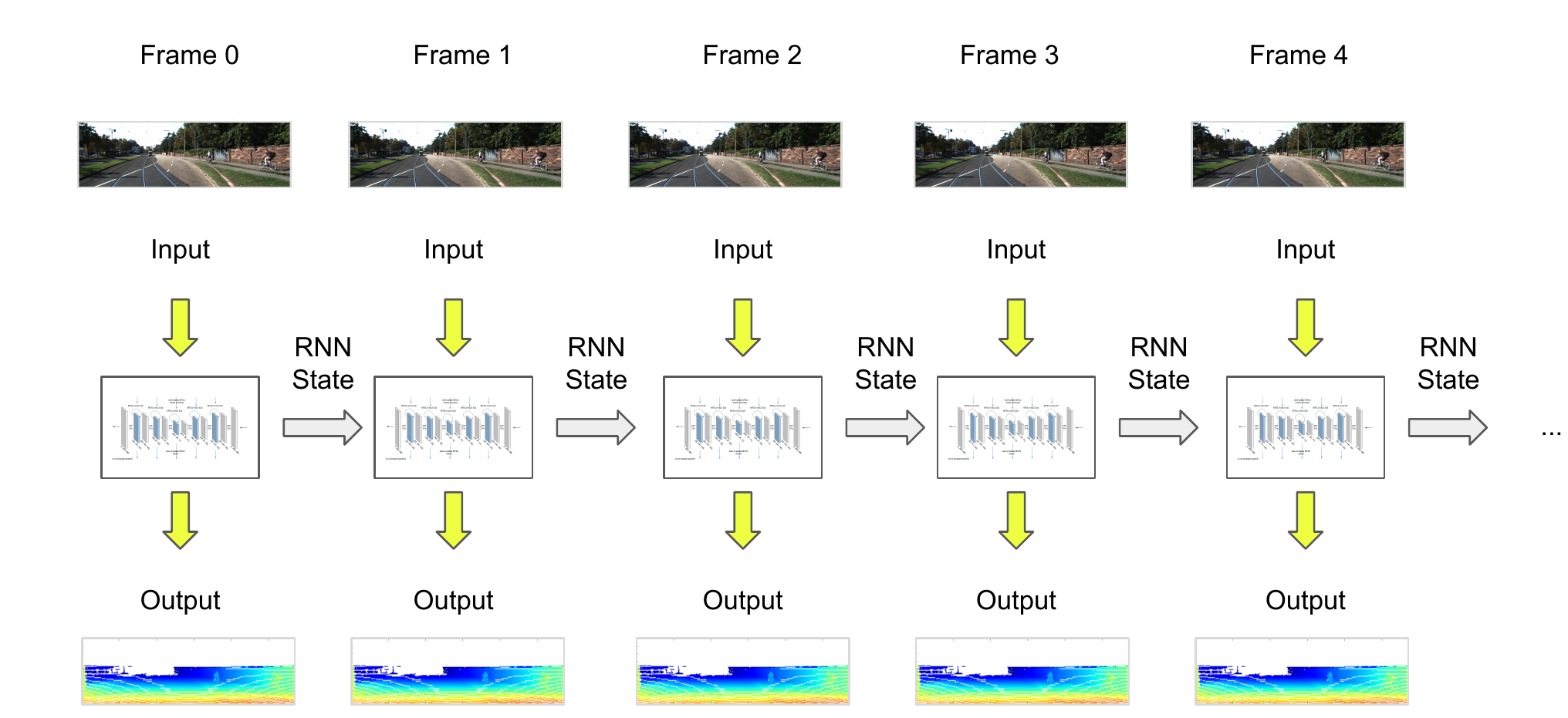}
\caption{Depth prediction model unrolled through time.  At each timestep, the recurrent neural network (RNN) receives one video frame and produces one depth prediction.  The states for the LSTM cells are updated at each timestep and the updated values are used for depth prediction on subsequent frames.  The output of the LSTM cells are passed to the next layer, while their states are passed through time to the next frame.}
\label{fig:unrolled_model}
\end{figure*}

\subsection{Depth Prediction from Monocular Video}

Figure~\ref{fig:model} shows the recurrent convolutional neural network that is used for predicting depth from monocular video.  Table~\ref{tab:arch} lists the architectural details.  The model consists of encoder and decoder parts, both with convolutional LSTM units. The convolutions in the encoder are with stride two without any max-pooling layers for downsizing the feature maps in the encoder.  Unlike tasks like object classification, the features in this domain are not invariant to translation.  So, we avoid using max-pooling layers to preserve the spatial structure of the feature maps.  In the decoder, the feature maps are gradually upsized to reach the input resolution.  Upsizing is done using depth-to-space layers~\cite{abadi2016tensorflow}, which spatially rearrange the activations, followed by convolutions.  The model uses convolutional LSTM cells at various spatial resolutions.  Convolutional LSTM cells are similar to regular LSTM cells~\cite{hochreiter2016long}, however, their gates are implemented by convolutions instead of fully-connected layers~\cite{shi2015convolutional}.  

Figure~\ref{fig:unrolled_model} shows the depth prediction model unrolled through time.  At each timestep, the network receives one video frame and produces one depth prediction.  Since the LSTM states are retained between subsequent frames, they enable the model to capture motion between two or more frames.  The output of the LSTM cells are passed to the next layer, while their states are passed through time to the next frame.  Therefore, the block processing frame $i$ receives the input frame $X_i$ and the LSTM states $S_{i-1}$ as inputs, where $S_i$ is the set of LSTM states from all layers after processing frame $i$, and $S_0 = 0$.  Unrolling the model simplifies training.  Although multiple copies of the network are instantiated, there is a single set of model parameters shared across the instances.  Our model applies layer normalization~\cite{ba2016layer} after each convolution or LSTM cell.  In recurrent networks layer normalization performs better than batch normalization.  

\begin{table}
\caption{Depth Prediction Model Architecture}
\begin{center}
\begin{tabular}{|l|l|c|}
 \hline
 Layer & Layer type and parameters & Activation size \\
 \hline
 Input & Input image & $88 \times 288 \times 3$ \\ 
 conv1 & Conv. $5 \times 5 \times 32$ stride 2 & $44 \times 144 \times 32$ \\
 conv-lstm1 & Conv-LSTM $5 \times 5 \times 32$ & $44 \times 144 \times 32$ \\
 conv2 & Conv. $3 \times 3 \times 64$ stride 2 & $22 \times 72 \times 64$ \\
 conv-lstm2 & Conv-LSTM $5 \times 5 \times 64$ & $22 \times 72 \times 64$ \\
 conv3 & Conv. $3 \times 3 \times 128$ stride 2 & $11 \times 36 \times 128$ \\
 conv-lstm3 & Conv-LSTM $5 \times 5 \times 128$ & $11 \times 36 \times 128$ \\
 ds1 & Depth-to-Space block size 2 & $22 \times 72 \times 32$ \\
 conv4 & Conv. $3 \times 3 \times 64$ stride 1 & $22 \times 72 \times 64$ \\
 conv-lstm4 & Conv-LSTM $5 \times 5 \times 64$ & $22 \times 72 \times 64$ \\
 ds2 & Depth-to-Space block size 2 & $44 \times 144 \times 16$ \\
 conv5 & Conv. $3 \times 3 \times 32$ stride 1 & $44 \times 144 \times 32$ \\
 conv-lstm5 & Conv-LSTM $5 \times 5 \times 32$ & $44 \times 144 \times 32$ \\
 ds3 & Depth-to-Space block size 2 & $88 \times 288 \times 8$ \\
 conv6 & Conv. $5 \times 5 \times 1$ stride 1 & $88 \times 288 \times 1$ \\
 sigmoid & Sigmoid & $88 \times 288 \times 1$ \\
 \hline 
\end{tabular}
\end{center}
\label{tab:arch}
\end{table}

We also experimented with more elaborate models, whose performance was not better than our model: 
adding skip connections; producing and consuming intermediate low-resolution predictions~\cite{fischer2015flownet};
adding a fully-connected layer plus dropout in the model bottleneck.

\subsection{Depth Prediction Loss Function}

We experimented with the $\mathcal{L}_2$ and reverse Huber losses.  

The $\mathcal{L}_2$ loss minimizes the squared Euclidean norm between predicted depth $D_i$ and ground truth depth label $Y_i$ for frame $i$: \mbox{$\mathcal{L}_2(D_i, Y_i) = \|D_i - Y_i\|^2_2$}.

The reverse Huber loss is defined in~\cite{laina2016deeper} as:

\begin{equation}
\mathcal{B}(\delta) {}=\displaystyle
   \begin{cases} 
      |\delta| & |\delta| \leq c, \\
      \frac{\delta^2 + c^2}{2c} & |\delta| > c
   \end{cases}
\end{equation}

\noindent where $\delta = D_i - Y_i$ and $c = \frac{1}{5} max_i(D_i^j - Y^j_i)$ where j iterates over all pixels in the depth map.  The reverse Huber loss computes the $\mathcal{L}_1$ norm when $|\delta| \leq c$ and the $\mathcal{L}_2$ norm otherwise.

Additionally, the loss equation can include an optional term to minimize the depth Gradient Difference Loss (GDL)~\cite{eigen2015predicting}, which is defined as:

\begin{align}
\begin{split}
\text{GDL}(D_i, Y_i) = \sum_{x,y} & \bigl|(D_i^{x,y} - D_i^{x-1,y}) - 
                                          (Y_i^{x,y} - Y_i^{x-1,y})\bigl|^2 + \\
                                  & \bigl|(D_i^{x,y} - D_i^{x,y-1}) -
                                          (Y_i^{x,y} - Y_i^{x,y-1})\bigl|^2
\end{split}
\end{align}

\noindent where $x, y$ iterate over pixel rows and columns in the depth map.  The purpose of the GDL term is to encourage local structural similarity between predicted and ground truth depth.

The final loss function is formed by computing the average loss over all frames in a sequence:

\begin{equation}
L(\theta) = \frac{1}{k} \sum_{i=1}^{k} \alpha_i L_\theta(D_i, Y_i)
\end{equation}

\noindent where $\theta$ represents all model parameters, $k$ is the number of frames in sequence, $\alpha_i$ is the scaling coefficient for frame $i$, and $L_\theta(D_i, Y_i)$ is equal to either $\mathcal{L}_2(D_i - Y_i) + \lambda_{\text{gdl}} \text{GDL}(D_i, Y_i)$ or $\mathcal{B}(D_i - Y_i) + \lambda_{\text{gdl}} \text{GDL}(D_i, Y_i)$.  In experiments we set $\alpha_i = 1$ for all $i$, and set $\lambda_{\text{gdl}}$ to either zero or one.  In all loss terms, we mask out pixels where there is no ground truth depth.

\begin{figure*}
\centering
\includegraphics[width=0.232\textwidth]{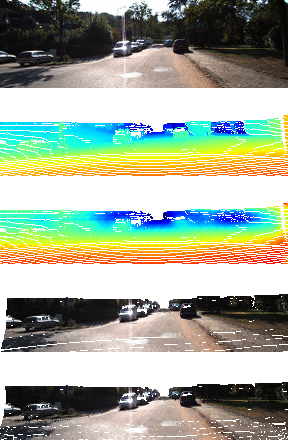}
\includegraphics[width=0.232\textwidth]{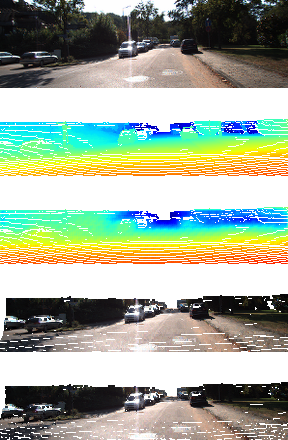}
\includegraphics[width=0.232\textwidth]{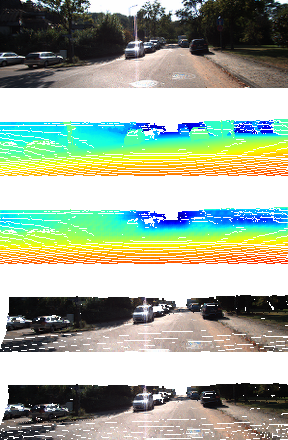}
\includegraphics[width=0.232\textwidth]{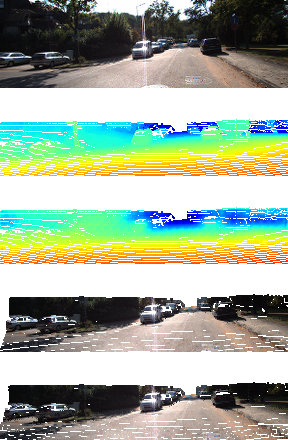}
\caption{Example depth and next frame predictions by our model from a four-frame sequence.  The four columns show the ground truth and predictions for frames 1-4.  From top to bottom in each column: 1) Input frame. 2) Ground truth depth 3) Predicted depth. 4) Next frame prediction constructed using ground truth depth 5) Next frame prediction constructed using predicted depth.  For frames 1-3 the ground truth next frame is visible at the top of the corresponding next column.  It can be seen that the quality of depth and next frame predictions improves as the model receives more video frames.  After seeing only the first frame, the model believes that the ground is closer than what it actually is (visualized by a stronger red hue.)  After seeing more frames, the depth estimate is improved.}
\label{fig:prediction_seq}
\end{figure*}

\subsection{Next Frame Prediction}

The next frame prediction is generated by additional transformation layers that are added after the depth output layer (not shown in figure~\ref{fig:unrolled_model}).  For each frame $i$, the next frame prediction $X'_i$ is generated using:

\begin{itemize}
  \item Video frame $X_{i-1}$ from the last timestep.
  \item Depth map prediction $D_{i-1}$ from the last timestep.
  \item Camera poses $P_{i-1}, P_i$.
\end{itemize}

First, the points in depth map $D_{i-1}$ are projected into a three-dimensional point cloud $C$.  The $x, y, z$ coordinates of the projected points in $C$ depend on their two-dimensional coordinates on the depth map $D_{i-1}$ as well as their depth values.  In addition to the three-dimensional coordinates, each point in $C$ is also assigned an RGB value.  The RGB value for each point is determined by its corresponding image pixel in $X_{i-1}$ located at the same image coordinates as the point's origin on depth map $D_{i-1}$.

Next, the camera's ego-motion between frames $i-1$ and $i$ are computed from pose vectors $P_{i-1}$ and $P_i$.  The computed ego-motion is six-dimensional and contains three translation components $t_x, t_y, t_z$ and three rotation components $r_x, r_y, r_z$.  Given the camera's new coordinates and principal axis, the point cloud $C$ is projected back onto a plane at a fixed distance from the camera and orthogonal to its principal axis.  Each projected point receives an updated depth value based on its newly-calculated distance to the projection plane.  The result of this projection is a depth map prediction $D_{i}$ for frame $i$.  Painting the projected points with their affixed RGB values creates the next frame prediction $X'_i$.  Embedding the matrix multiplications that represent the necessary projections and transformations in the model allows it to directly produce next frame predictions.

The net effect of the two projections and the intermediate translation and rotation is to move pixels in $X_{i-1}$ to updated coordinates in the predicted image $X'_i$.  The magnitude and direction of the movement for each pixel is a function of the depth of the pixel's corresponding point in the depth maps, and the magnitude and direction of ego-motion components.

Since different pixels may move by different amounts, this process can produce overlapping pixels as well as gaps where no pixel moves to a given coordinate in the next frame prediction.  The overlaps are resolved by picking the point whose depth value is smaller (closer to the camera).  In our implementation the gaps are partly filled using a simple splatting technique which writes each point over all four image pixels that it touches.  A more sophisticated approach based on inpainting~\cite{chen2013knn} can be used to fill in all the gaps.

\section{EXPERIMENTAL EVALUATION}

We test our approach on the KITTI dataset~\cite{geiger2013vision} which is collected from a vehicle moving in urban environments.  The vehicle is equipped with cameras, lidar, GPS, and inertial sensors.  The dataset is organized as a series of videos with frame counts ranging from about 100 to a few thousands.  For each frame, the dataset contains RGB images, 3D point clouds, and the vehicle's pose as latitude, longitude, elevation, and yaw, pitch, roll angles.
We split the videos into training and evaluation sets and generate 10-frame sequences from each video.  In total, our dataset contains about 38000 training sequences and 4200 validation sequences.

\begin{figure}
\centering
\hspace{-0.35cm}\includegraphics[width=0.253\textwidth]{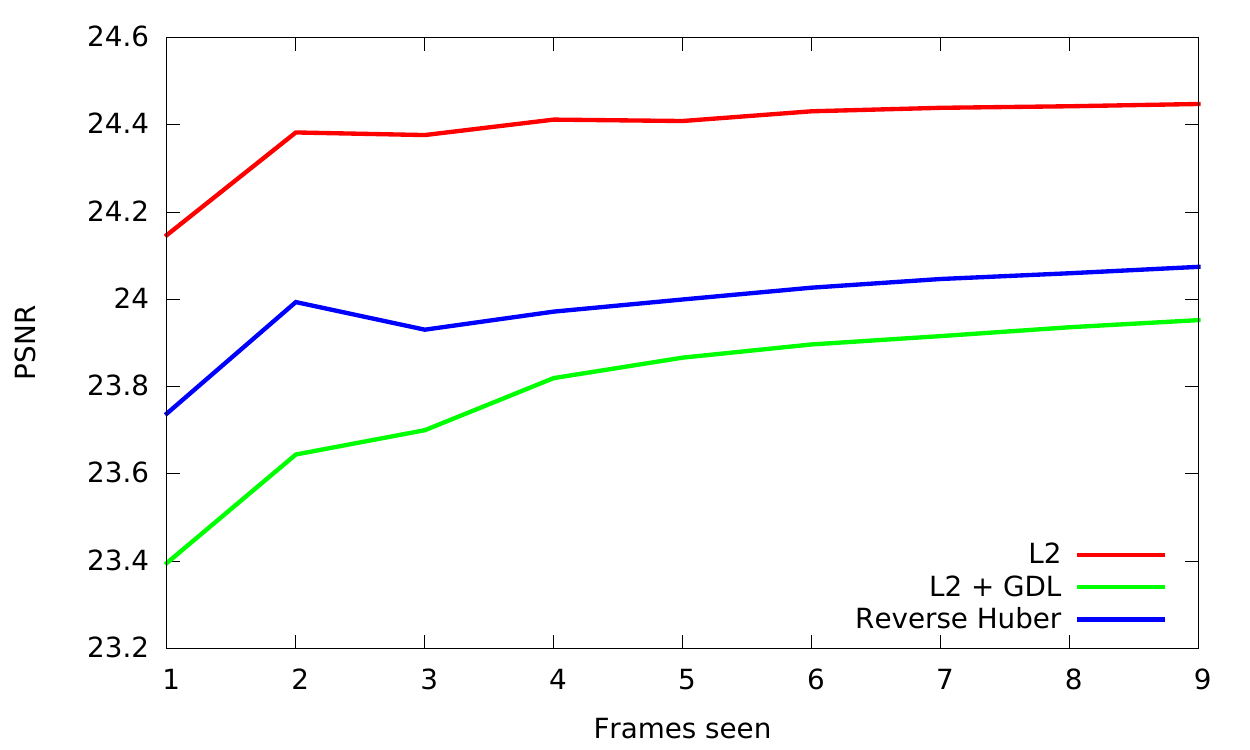}
\hspace{-0.15cm}\includegraphics[width=0.253\textwidth]{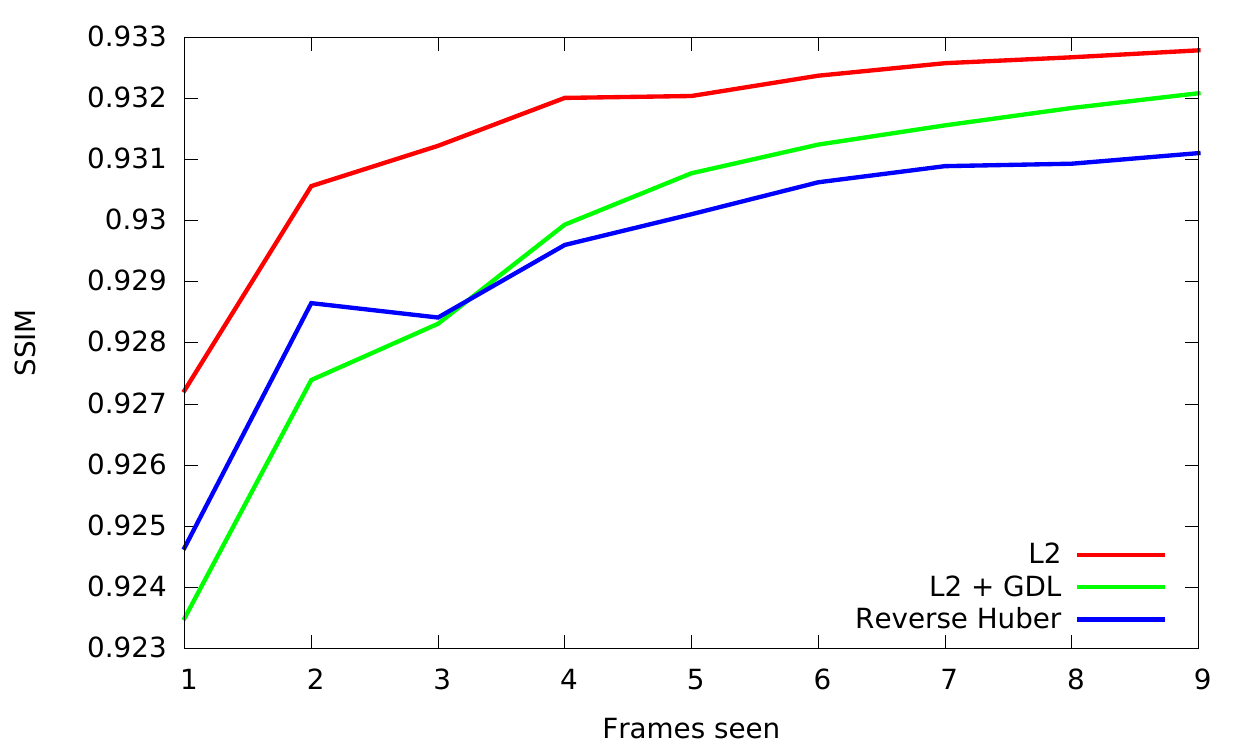}
\caption{Next frame prediction quality metrics as a function of the number of frames seen.  The plot shows PSNR and SSIM metrics for different loss functions.  For all loss functions, our model performs better as it receives more video frames.  The biggest jump in quality happens between frames one and two, since at that point motion information becomes available to the model.  PSNR improves moderately thereafter, whereas SSIM continues to improve as more video frames are used.  
}
\label{fig:metrics_by_frame}
\end{figure}

\begin{figure*}[ht]
\centering
\includegraphics[width=0.9\textwidth]{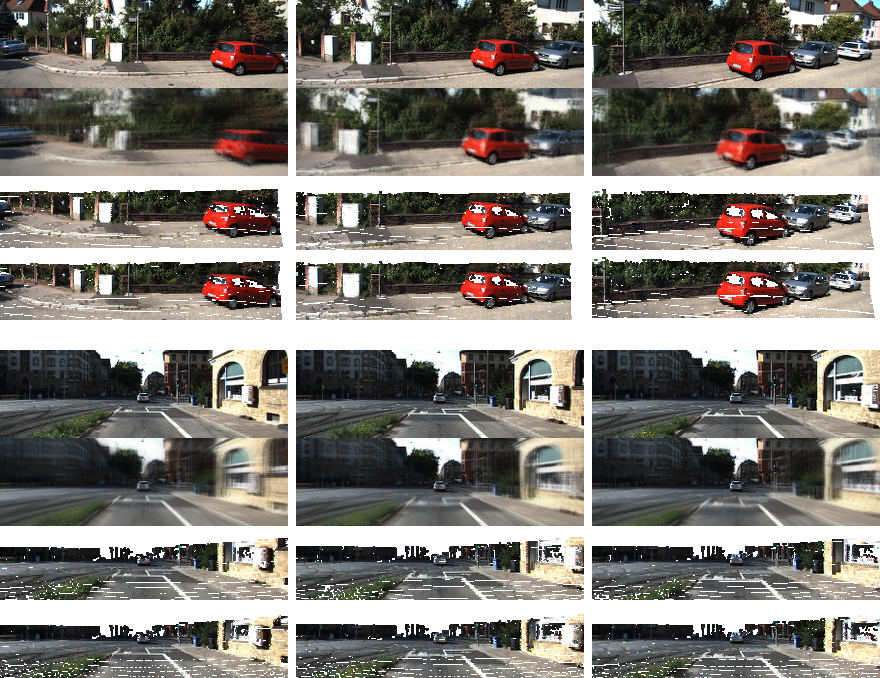}
\caption{
Comparison of next frame prediction by our method with state-of-the-art video prediction model STP~\cite{finn2016unsupervised}.  The frames shown are from the first sequence in two validation videos.  In total six ground truth frames from the two videos are shown.  From top to bottom in each column: 1) Input last frame. 2) Prediction by STP~\cite{finn2016unsupervised}. 3) Next frame prediction using ground truth depth. 4) Next frame prediction using predicted depth by our method.  While the STP model can predict the location of objects well, it produces blurry images, especially when there is large motion.}
\label{fig:compare_with_dna}
\end{figure*}

\begin{figure}[!ht]
\vspace{-0.2cm}
\includegraphics[width=0.45\textwidth,trim={-2.0cm 6.2cm 0 6.6cm},clip]{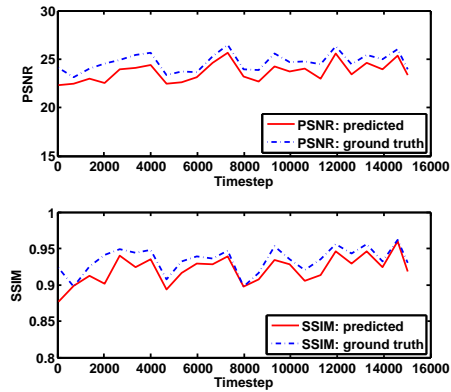}
\caption{
Comparing quality of next frame prediction from our method using predicted depth vs next frame predictions using ground truth depth.  Top: PSNR.  Bottom: SSIM.  Metrics are measured over small samples as training progresses.  Both SSIM and PSNR are closely tracking the best predictions possible under our method.}
\label{fig:compare_with_gt}
\end{figure}

\subsection{Generating Ground Truth Depth Maps}

We generate ground truth depth maps by first transforming the point clouds using the calibration matrices in the KITTI dataset and then projecting the points onto a plane at a fixed distance from the camera. The points that are included in the depth map have depth values ranging from 3m (approximate cutoff for points visible by camera) to 80m (sensor's maximum range.)  Instead of requiring the model to predict such large values, we use $(3.0 \mathbin{/} \text{depth})$ as labels.  We also experimented with using $\log(\text{depth})$ as labels.  Additionally, the labels are linearly scaled to values in the interval $[0.25, 0.75]$.  This normalization helps reduce the imbalance between the importance of accurate predictions for points that are far and near.  Without normalization, loss terms like $\mathcal{L}_2$ can give disproportionate weights to near and far points.

The generated depth maps contain areas with missing depth due to a number of causes:  1) Since the camera and the lidar are at different locations on the car, there are often overlaps and shadows when the point cloud is viewed from the camera.  2) Objects that are farther than the sensor's range (80m) and objects that do not reflect the light back to the sensor (shiny objects) are not detected by the sensor.  3) Since the point clouds are sparse, there are image coordinates that usually do not line up with any point.

\subsection{Quality Metrics}

We employ two image quality metrics~\cite{hore2010image} to evaluate next frame predictions produced by the model:

\begin{itemize}
\item Peak signal-to-noise ratio (PSNR)
\item Structural similarity (SSIM)
\end{itemize}

Both metrics are standard in measuring the quality of image predictions~\cite{oh2015action,mathieu2015deep,finn2016unsupervised}.  Using these metrics allows us to measure the quality of predictions independently of the loss functions and depth transform functions used.  The metrics are computed for pixels where the model makes a prediction.

\subsection{Training}

Our model is implemented in TensorFlow~\cite{abadi2016tensorflow}.  We use the Adam optimizer~\cite{kingma2014adam} with a learning rate of $0.0001$.  The model weights are initialized from a Gaussian distribution with a standard deviation of $0.01$.  The LSTM states are initialized to 0.0 and the forget gates are initialized to 1.0.  Each training timestep processed a mini-batch of eight 10-frame sequences. We use the $\mathcal{L}_2$ loss for training.

\begin{figure}
\vspace{.3cm}
\hspace{-1.3cm}
\includegraphics[width=0.6\textwidth, trim={0 6.9cm 0 6.7cm},clip]{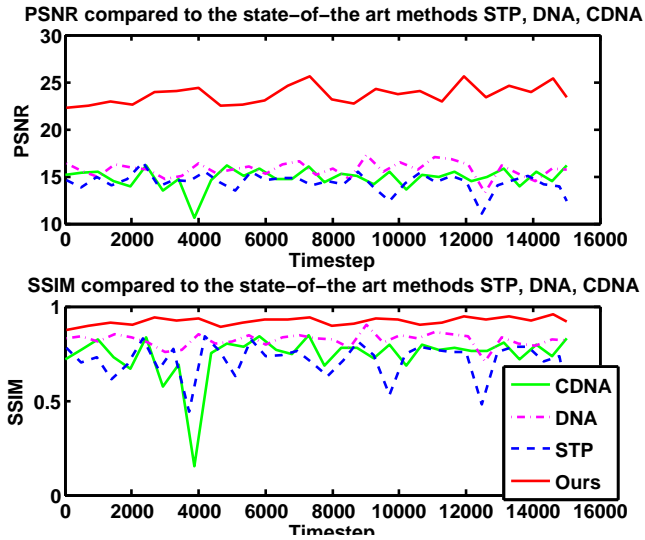}
\caption{Comparison of quality metrics on next frame predictions by our model against STP, DNA, and CDNA methods~\cite{finn2016unsupervised} over the validation dataset (higher values are better). Our model clearly outperforms the previous ones in both metrics.  For our model, the metrics are computed on areas where the model produces predictions.  The fluctuations are partly due to the relatively small sample sizes used in the evaluation process.  Each point represents an average over 72 next frame predictions from eight random sequences.}
\label{fig:compare_metrics}
\end{figure}

\subsection{Results}

Figure~\ref{fig:prediction_seq} shows the outputs generated by our trained model.  The figure shows depth predictions and next frame predictions from four frames of a sequence.  Each column corresponds to one frame.  The first row shows the ground truth last frame and the last row shows next frame predictions generated using predicted depth.
By comparing the quality of depth predictions for each frame it can be observed that the model's predictions are improving with seeing more frames.  After seeing only the first frame, the model believes that the ground is closer than what it actually is.  By receiving more input images, the model's depth prediction improves and turns more similar to ground truth.

This observation is also supported quantitatively in Figure~\ref{fig:metrics_by_frame}, which shows how the quality metrics improve as a function of the number of prior frames seen by the model. The plot shows per-frame PSNR and SSIM averages over 100 mini-batches for different loss functions.  As seen, for all loss functions and both metrics, the model performs better as it receives more video frames.  The biggest jump in quality occurs between frames one and two, when the model gains access to the motion information in the sequence.  However, the metrics continue to improve with more frames.

We compare our next frame predictions using predicted depth, to predictions generated when using ground truth depth. The last two rows in Figure~\ref{fig:prediction_seq} show a comparison in visual quality.
Figure~\ref{fig:compare_with_gt} plots the quality difference between these two sets of predictions using SSIM and PSNR metrics.  These plots show that our model's predictions track closely the best predictions possible that are based on known depth.

\subsection{Comparison to State of the Art}

We compare our model with the state-of-the-art video prediction models~\cite{finn2016unsupervised}, which are the best and most recent models for next frame prediction. We trained three model variants DNA, CDNA, and STP on our dataset.  All these models are action-conditioned \ie[,] they have placeholders to receive state and action inputs.  In addition to the video images, we pass the current camera pose as the state and the computed ego-motion as the action to all three models.

Figure~\ref{fig:compare_with_dna} qualitatively compares the next frame predictions generated by our model and by the prior methods~\cite{finn2016unsupervised}. As we can see,~\cite{finn2016unsupervised} is usually able to place the objects at the right location in the next frame.  However, it produces fuzzy predictions, especially when the scene background is moving between frames.  This is due to using a loss function based on RGB values in the next frame, which causes the network to predict a weighted average of potential outcomes.  Our method, on the other hand, produces sharp and accurate predictions.

Figure~\ref{fig:compare_metrics} quantitatively compares the predictions generated by our model with DNA, CDNA, and STP.  The predictions by our method outperform prior models on both average PSNR and SSIM metrics.  In terms of PSNR, our model performs much better by producing results in the order of 24-25, whereas the three prior methods from~\cite{finn2016unsupervised} produce values in range 15-17.  Similarly for SSIM, with a maximum possible value of 1.0, our model achieves ~0.92-0.93, whereas for prior methods the value is around 0.7-0.8.

As seen in the results, our method produces higher quality images.  Existing next-frame prediction methods that we know of generate images with different levels of blur, which is caused by the inherent uncertainty imposed by the unstructured loss function.  The gaps in the output generated by our model are due to the resolution of depth map predictions and the geometric transformations.  They can be inpainted~\cite{chen2013knn} if better visual quality is desired.

\begin{figure}[!ht]
\centering
\includegraphics[width=0.493\linewidth]{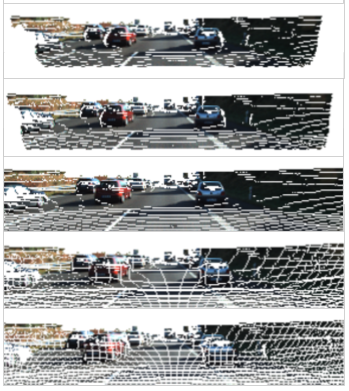}
\includegraphics[width=0.493\linewidth]{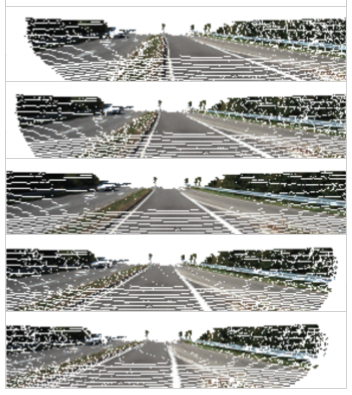}
\caption{Next frame simulations using ground truth depth and hypothetical ego-motions. Middle row: current frame.  Other rows: Simulated next frames for moving forward/backward (left) and for moving sideways (right).  See video at \texttt{https://goo.gl/gvJvZC}.
}
\label{fig:egomotion1}
\end{figure}

\vspace{-0.1cm}
\subsection{Failure cases}

We have observed cases where the depth of thin elongated objects \eg[,] poles are not estimated correctly. Since our approach is based on depth estimation, this affects the quality of our next frame predictions.  The primary reason behind these errors is probably the low impact of these objects in the loss equation.  Another contributing factor is the imperfect alignment of depth maps and video frames in the training data, which affects thin objects more.  These misalignments are primarily due to varying time delays between the rotating lidar and the camera for different regions of the image.  

\subsection{Simulating Hypothetical Next Frames}

Our approach can be used to generate potential next frames based on hypothetical ego-motions of the moving agent e.g. the moving vehicle. Such next frame simulations can be useful for exploring the scene in 3D, given a hypothetical motion; they are not intended to model individual motions of dynamic objects in the scene. 
Figure~\ref{fig:egomotion1} shows example next frame simulations based on a range of hypothetical ego-motions corresponding to moving forward/backward and sideways. The frames shown are generated using ground truth depth.  
These results are best viewed as an animation.  Please see the accompanying video at \texttt{https://goo.gl/gvJvZC}.

\section{Conclusions and Future work}

We present a new method for predicting the next frame from monocular video using the scene geometry.  Our method uses an RCNN that is trained to predict depth from a sequence of images.  The experiments show that our model can capture the motion between subsequent frames and improve its depth predictions.  Compared to existing work, our model produces higher quality next frame predictions.

We can improve the visual quality of the predictions by upsampling and inpainting ground truth depth maps and inpainting next frame predictions where possible.  Predicting multiple frames into the future is a natural extension to this work.  Similarly, the camera's near-future trajectory can be predicted from prior observations.  Predicting the depth map for future frames instead of the input frame would allow for capturing the motion of dynamic objects in the scene.
Applying our approach to anomaly detection will be an important next step.  For example, we can superimpose our next frame prediction with the actually observed frame and analyze the mismatches in the scene topology (depth) or appearance (RGB frame).  Large mismatches may be an indication of an object moving with an unexpected velocity, and can be used as informing signals for safer navigation.

\section*{Acknowledgments}

We thank Chelsea Finn for sharing the code for their models and the Google Brain team for discussions and support.

\bibliographystyle{plain}
\bibliography{refs.bib}

\end{document}